# Explicit Cognitive Allocation: A Principle for Governed and Auditable Inference in Large Language Models.

*Héctor Manuel Manzanilla-Granados, **Zaira Navarrete-Cazales. *Miriam Pescador-Rojas, *Tonahtiu Ramírez-Romero.

*Escuela Superior de Cómputo del I.P.N, Av. Juan de Dios Bátiz s/n esq. Av. Miguel Othón de Mendizabal. Colonia Lindavista. Alcaldia: Gustavo A. Madero. C. P. 07738. Ciudad de México.

**Facultad de Filosofía y Letras, Circuito Interior s/n, Ciudad Universitaria, Coyoacán, C.P. 04510, Ciudad de México.

Fecha: 15/12/2025

hmanzanilla@ipn.mx

**Abstract.**

The rapid adoption of large language models (LLMs) has enabled new forms of AI-assisted reasoning across scientific, technical, and organizational domains. However, prevailing modes of LLM use remain cognitively unstructured: problem framing, knowledge exploration, retrieval, methodological awareness, and explanation are typically collapsed into a single generative process. This cognitive collapse limits traceability, weakens epistemic control, and undermines reproducibility, particularly in high-responsibility settings.

We introduce Explicit Cognitive Allocation, a general principle for structuring AI-assisted inference through the explicit separation and orchestration of epistemic functions. We instantiate this principle in the Cognitive Universal Agent (CUA), an architecture that organizes inference into distinct stages of exploration and framing, epistemic anchoring, instrumental and methodological mapping, and interpretive synthesis. Central to this framework is the notion of Universal Cognitive Instruments (UCIs), which formalize the heterogeneous means—computational, experimental, organizational, regulatory, and educational—through which abstract inquiries become investigable.

We evaluate the effects of explicit cognitive and instrumental allocation through controlled comparisons between CUA-orchestrated inference and baseline LLM inference under matched execution conditions. Across multiple prompts in the agricultural domain, CUA inference exhibits earlier and structurally governed epistemic convergence, substantially higher epistemic alignment under semantic expansion, and systematic exposure of the instrumental landscape of inquiry. In contrast, baseline LLM inference shows greater variability in alignment and fails to explicitly surface instrumental structure.

Together, these results establish explicit cognitive and instrumental allocation as a model-agnostic architectural principle for improving the controllability, auditability, and epistemic scope of AI-assisted reasoning. The proposed metrics and workflow provide a foundation for the systematic evaluation of reasoning architectures in domains where transparency, governance, and human oversight are critical.

**Introduction.**

Artificial intelligence has become deeply embedded in contemporary scientific, technical, and organizational practice. Large language models (LLMs) and foundation models have reshaped how problems are framed, how prior knowledge is accessed, and how explanatory narratives are constructed across domains ranging from molecular biology and medicine to software engineering and the social sciences[1–4]. At the same time, an expanding ecosystem of computational tools and infrastructures has amplified the analytical capacity available to researchers, further accelerating the integration of AI into epistemic workflows.

Recent research has focused on improving the reasoning capabilities of LLMs by introducing structured prompting strategies and agentic frameworks. Approaches such as chain-of-thought prompting explicitly expose intermediate reasoning steps to improve performance on complex tasks[5]. Subsequent work has explored alternative organizational structures for reasoning, including tree-based and graph-based formulations, demonstrating that the structure of deliberation can substantially affect coherence and problem-solving behavior[6,7]. Parallel efforts have investigated how reasoning traces can be evaluated, controlled, or selectively suppressed to balance interpretability, efficiency, and performance[8].

Despite these advances, the prevailing paradigm remains largely model-centric. Reasoning, retrieval, instrumental suggestion, planning, and explanation are typically treated as emergent properties of a single generative process. Even retrieval-augmented generation systems, which integrate external knowledge sources into inference, primarily emphasize architectural routing and data access rather than explicit separation of epistemic roles[9]. As a result, complex inquiries are often addressed through conversational prompting that conflates conceptual framing, epistemic grounding, methodological awareness, and interpretive synthesis into a single interactional flow.

This functional conflation has important consequences. Agent-based and tool-augmented systems increasingly emphasize autonomy, execution, and adaptive supervision, particularly in applied settings[10]. While such systems can be operationally effective, they often obscure the epistemic structure of reasoning itself, making it difficult to trace how conclusions emerge, which assumptions are invoked, or which alternative instrumental pathways were considered but discarded. These limitations are especially consequential in high-responsibility domains where transparency, accountability, and human oversight are critical[11].

We argue that addressing these limitations requires moving beyond model-centric intelligence toward explicit cognitive organization. Drawing on theories of distributed cognition and social epistemology, scientific reasoning is understood here as a coordinated process involving heterogeneous cognitive resources—human judgment, institutional practices, methodological instruments, and material infrastructures—rather than as the output of an isolated agent[12–14]. Within this perspective, artificial intelligence is best treated not as an autonomous epistemic authority, but as a mediator that supports distinct phases of inquiry under explicit governance.

In this work, we introduce Explicit Cognitive Allocation, a general principle for structuring AI-assisted reasoning through the deliberate separation and orchestration of epistemic functions. We instantiate this principle in the Cognitive Universal Agent (CUA), an architecture that organizes inference into non-executive cognitive stages dedicated to conceptual framing, epistemic anchoring, instrumental and methodological mapping, and interpretive synthesis. Rather than executing tools or making decisions, the CUA renders the epistemic and instrumental structure of inquiry explicit, preserving full human authority over execution, validation, and action.

We further introduce a set of operational metrics to evaluate how explicit cognitive allocation affects inference behavior, including convergence discipline, semantic alignment under expansion, and instrumental visibility. Through controlled comparisons with baseline LLM inference, we demonstrate that explicit cognitive and instrumental organization improves the traceability, stability, and epistemic scope of AI-assisted reasoning. Together, these contributions establish explicit cognitive allocation as a model-agnostic architectural principle for developing AI systems that support transparent, governed, and human-centered inquiry.

## 2. Explicit Cognitive Allocation

Explicit Cognitive Allocation is a design principle for AI-assisted reasoning systems that governs how cognitive labor is structured, separated, and coordinated during inference. Rather than treating reasoning as an undifferentiated generative process, this principle requires that distinct epistemic functions—such as conceptual framing, epistemic grounding, instrumental awareness, and interpretive synthesis—be made explicit, staged, and externally legible throughout the reasoning workflow.

At its core, Explicit Cognitive Allocation asserts that the quality, controllability, and accountability of AI-assisted reasoning depend not primarily on model capacity, prompt sophistication, or execution autonomy, but on the organization of cognitive roles within the inference process. The principle shifts the focus of system design from optimizing output generation to governing how inquiry unfolds: how an initial human intent is progressively transformed into structured epistemic artifacts under explicit constraints.

Under Explicit Cognitive Allocation, inference is conceived as a sequence of epistemic transformations rather than as a single act of generation. Each transformation serves a distinct cognitive purpose and produces an intermediate artifact that stabilizes and constrains subsequent reasoning. These artifacts—such as structured problem framings, epistemic anchors, instrumental maps, and integrated syntheses—are not incidental by-products of prompting, but first-class objects of the reasoning process. Their explicit production enables traceability, reproducibility, and human inspection of how conclusions emerge.

Crucially, Explicit Cognitive Allocation is agnostic to the underlying language model. It does not prescribe architectural changes to transformers, training regimes, or decoding strategies. Instead, it operates at the level of inference organization, specifying how model invocations are coordinated and what epistemic role each invocation is permitted to play. This model-agnostic character allows the principle to be applied across diverse LLMs and deployment contexts, including scientific analysis, technical planning, policy exploration, and organizational decision support.

A defining feature of Explicit Cognitive Allocation is the separation between epistemic reasoning and operational execution. Reasoning stages are designed to identify, contextualize, and compare possible instruments of inquiry—computational methods, experimental paradigms, institutional procedures, regulatory frameworks—without executing them. This separation preserves human authority over action while ensuring that the space of available options, constraints, and trade-offs is made visible prior to decision-making. In this sense, Explicit Cognitive Allocation supports governed inference: reasoning processes whose structure is externally constrained, auditable, and resistant to uncontrolled semantic drift.

Existing approaches to structured reasoning in large language models implicitly gesture toward this need but stop short of enforcing it. Chain-of-thought prompting, tree-based reasoning, graph-based representations, and agentic tool-use frameworks introduce internal structure within a single generative loop, often combining exploration, evaluation, planning, and execution in tightly coupled sequences. While these approaches can improve task performance or coherence, they typically allow cognitive roles to intermingle dynamically, with little external visibility into which epistemic function is being performed at a given moment or how instrumental alternatives are surfaced and constrained.

As a result, reasoning trajectories frequently remain opaque. It becomes difficult to determine whether a conclusion reflects careful epistemic anchoring, opportunistic pattern completion, or premature instrumental fixation. In high-responsibility domains—such as scientific discovery, public policy, infrastructure planning, or environmental governance—this opacity undermines accountability and limits the usefulness of AI-assisted reasoning as a decision-support tool rather than an automated answer generator.

Explicit Cognitive Allocation addresses this limitation by externalizing cognitive structure. Instead of relying on emergent internal coherence, it enforces a disciplined division of epistemic labor, ensuring that exploration precedes grounding, that instrumental awareness precedes methodological commitment, and that synthesis occurs only after the relevant epistemic landscape has been rendered explicit. This organization transforms reasoning from a conversational artifact into a governed epistemic process.

Within this framework, efficiency is redefined. Rather than minimizing token usage or inference latency, efficiency is understood as epistemic efficiency: the ability to reach stable, well-aligned, and instrumentally informed convergence with minimal semantic drift and maximal transparency. As demonstrated in the evaluation presented below, explicitly allocated workflows often converge in fewer epistemically meaningful steps, even when they incur higher raw token counts. The additional computational cost reflects not redundancy, but the deliberate externalization of structure required for auditability and control.

Explicit Cognitive Allocation thus provides a principled foundation for the design of AI-assisted reasoning systems that prioritize transparency, traceability, and human governance over autonomous execution. It reframes intelligence not as the capacity to act, but as the capacity to organize inquiry in a way that supports responsible human judgment. In the following sections, we operationalize this principle through a functional taxonomy of AI cognitive roles and instantiate it in the Cognitive Universal Agent (CUA), an architecture explicitly designed to enforce governed, auditable inference.

## 3. Cognitive Universal Agent (CUA): High-level Architecture

Explicit Cognitive Allocation (ECA) provides a general principle for structuring AI-assisted reasoning, but it does not, by itself, specify how such allocation is realized within an inference system. To instantiate this principle in a concrete, reproducible, and model-agnostic manner, we introduce the Cognitive Universal Agent (CUA): an architectural framework that governs how epistemic functions are separated, orchestrated, and constrained during AI-assisted inquiry.

The CUA is not a task-specific agent, an autonomous decision-maker, or an execution-oriented system. Rather, it functions as a *mediational cognitive architecture* that structures how reasoning unfolds across distinct epistemic stages. Its contribution lies not in performing computation, optimization, or action, but in ensuring that exploration, epistemic anchoring, instrumental awareness, and synthesis are not collapsed into a single generative operation. In doing so, the CUA renders reasoning processes transparent, traceable, and interpretable for human judgment.

### 3.1 Architectural role and design principles

At a high level, the CUA operates as a governing layer over large language models and retrieval mechanisms, imposing epistemic discipline through externally defined roles, stage boundaries, and explicit non-executive constraints. Unlike conventional LLM-based systems, in which framing, retrieval, reasoning, and explanation emerge implicitly from a single generative pass, the CUA enforces an explicit division of cognitive labor aligned with the principle of Explicit Cognitive Allocation.

Three architectural commitments define the CUA:

1. **Functional separation.** Distinct epistemic roles are assigned to different stages of inference, preventing the conflation of conceptual exploration, knowledge grounding, methodological awareness, and interpretive synthesis.

2. **Non-executive operation.** The CUA does not perform numerical computation, simulation, experimentation, or external action. All operations are epistemic and organizational, preserving human authority over execution and validation.

3. **Traceability and auditability.** Each stage of inference produces structured epistemic artifacts that can be inspected, logged, and evaluated independently of the final narrative output.

Together, these commitments ensure that the CUA governs *how* reasoning proceeds without dictating *what* conclusions should be drawn.

### 3.2 Cognitive functions and epistemic roles

The architecture of the CUA is organized around a small set of functionally differentiated cognitive roles, each defined not by algorithms or model internals, but by the epistemic work it performs within inquiry. At a conceptual level, the CUA distinguishes between functions responsible for exploration and framing of ill-defined human intentions; epistemic anchoring within validated bodies of knowledge and stabilized practice; instrumental and methodological mapping of available means of investigation; and interpretive synthesis of heterogeneous epistemic materials.

These roles correspond to generative, retrieval-oriented, planning-oriented, simulation-oriented, and explanatory functions, respectively. Their precise instantiation, interaction, and constraints are described in detail in the Supplementary Information. In the main text, it is sufficient to emphasize that no single role is permitted to dominate or subsume the others. The architecture is explicitly designed to prevent the cognitive collapse characteristic of monolithic generative inference, in which framing, retrieval, reasoning, and explanation are implicitly entangled within a single model invocation.

By enforcing role separation, the CUA transforms AI systems from opaque generative engines into components of a governed epistemic process.

**3.3 The AI-assisted workflow**

The epistemic roles defined above are orchestrated through a structured *AI-assisted workflow* that specifies the order, scope, and constraints under which cognitive functions may operate. The workflow is sequential but not rigid: it enforces epistemic dependencies without prescribing specific content, methods, or outcomes.

Conceptually, the workflow proceeds through four stages: exploration and framing, where the inquiry is clarified and alternative conceptual interpretations are surfaced; epistemic anchoring, where relevant prior knowledge, theoretical frameworks, and stabilized practices are identified; instrumental and methodological mapping, where available means of investigation are organized and contextualized without triggering execution; and interpretation and synthesis, where the outputs of prior stages are integrated into a coherent epistemic representation.

A central feature of this workflow is the explicit separation between epistemic convergence and narrative realization. Convergence—defined as the closure of epistemic commitments—occurs before any final narrative rendering intended for human readability. This separation allows the system to distinguish between reasoning that establishes epistemic structure and generation that merely expresses that structure linguistically.

**3.4 What the CUA is—and is not**

The Cognitive Universal Agent is deliberately conservative in scope. It is not an autonomous agent, a planner that executes actions, or a system that claims empirical discovery. It does not optimize objectives, select tools, validate hypotheses, or perform experiments. Instead, it operates as a cognitive scaffold that supports disciplined reasoning by making assumptions, constraints, and available instruments explicit.

At the same time, the CUA is not domain-specific. Its universality derives from invariants of scientific and technical reasoning—explicit framing, epistemic grounding, instrumental mediation, and interpretive synthesis—rather than from embedded domain knowledge or specialized training. As a result, the same architectural principles apply across scientific, industrial, organizational, and creative contexts.

**3.5 Architectural implications**

By externalizing cognitive organization into an explicit architecture, the CUA transforms large language models from monolithic generators into components of a governed epistemic workflow. This shift has two immediate implications. First, it enables reproducible and auditable reasoning paths that can be evaluated independently of output fluency or stylistic quality. Second, it provides a principled foundation for comparing AI-assisted reasoning systems based on how they allocate cognitive labor, rather than on surface-level answer quality alone.

In the following sections, we build on this architecture to examine how explicit cognitive allocation supports systematic instrumental awareness and to evaluate its effects through controlled comparisons with baseline LLM-based inference.

**4. Universal Cognitive Instruments (UCIs) and the Operational Substrate of Inquiry**

Scientific reasoning—whether computational, experimental, organizational, social, or creative—is never performed in isolation. It is mediated by instruments: structured epistemic resources that externalize expertise, stabilize procedures, encode methodological constraints, and render complex problems tractable. These instruments shape how inquiry unfolds by enabling certain forms of investigation while constraining others, thereby determining what can be asked, what can be measured, and what can be justified.

A crucial implication follows: not all instruments are computational, and even when computation is involved, not all instruments are directly executable by artificial systems. Much of what renders inquiry reliable, accountable, and socially legitimate—experimental protocols, methodological standards, institutional procedures, regulatory frameworks, ethical oversight mechanisms, training practices, and geographically or contextually grounded constraints—exists outside software. These instruments require human judgment, institutional responsibility, and contextual interpretation. AI-assisted frameworks that equate problem solving with autonomous tool execution therefore risk misrepresenting how real-world investigation and decision-making actually occur.

Within the Cognitive Universal Agent (CUA), we introduce the concept of Universal Cognitive Instruments (UCIs) to denote the heterogeneous set of epistemic resources through which humans systematically investigate and address complex problems across domains. UCIs include computational methods and analytical frameworks, but also encompass experimental methodologies, technical documentation, conceptual frameworks, institutional and regulatory procedures, organizational practices, and educational resources. What unifies these instruments is not their technological implementation, executability, or disciplinary origin, but their epistemic role as mediators of inquiry: they structure how knowledge is produced, situated, communicated, and rendered actionable.

Crucially, the CUA does not execute these instruments. It does not run software, perform numerical computation, conduct experiments, file permits, or enact institutional procedures. Its role is strictly epistemic and organizational: to identify, map, and contextualize the instruments that are relevant to a given epistemic intent. In doing so, the CUA produces a structured representation of the available cognitive resource space—what instruments exist, how they are typically used, what assumptions they encode, and what constraints they

impose—thereby supporting informed human judgment without collapsing reasoning into action.

This distinction is foundational. In the CUA framework, instruments are not actions performed by the agent, but resources identified and organized by the agent. Execution, experimentation, and institutional action remain explicitly outside the CUA and under human control, while the CUA operates exclusively at the level of epistemic organization and methodological awareness.

**4.1 Universal Cognitive Instruments (UCIs)**

Universal Cognitive Instruments (UCIs) denote the heterogeneous set of structured epistemic resources—computational, experimental, methodological, conceptual, institutional, organizational, regulatory, geographical, economic, ethical, and educational—through which abstract questions are connected to feasible paths of investigation. UCIs are defined not by their executability or technological form, but by their epistemic function: they externalize expert knowledge, stabilize procedures, encode accumulated methodological constraints, and structure how inquiry is conducted and interpreted.

UCIs simultaneously enable and constrain inquiry. These constraints are not incidental limitations but constitutive features of the epistemic landscape in which reasoning unfolds. Regulatory prohibitions, institutional mandates, data availability, infrastructural limitations, or ethical boundaries may render technically plausible approaches methodologically or socially inadmissible. Within the CUA, such instrumental constraints and tensions are treated as first-class epistemic structures: they are rendered explicit and preserved, rather than resolved, optimized, or adjudicated by the system.

The term *universal* is used here in a strictly procedural sense. UCIs are not universal because the same instruments apply across all problems, but because the process by which relevant instruments are identified, organized, and contextualized can be made domain-invariant. Across scientific, technical, organizational, social, and creative inquiries, the same recurring question applies: *which instruments exist for problems of this type, and what do they enable or constrain?* While the specific instruments vary by domain, scale, geography, and institutional context, the logic by which they are mapped remains stable.

**4.2 Computational Methodological Instruments (CMIs): Positioning and Scope.**

Within the broader space of Universal Cognitive Instruments, Computational Methodological Instruments (CMIs) constitute a specific subclass corresponding to instruments whose function is to carry out formally specified computational operations within explicit technical frameworks. CMIs include numerical solvers, optimization libraries, simulation environments, data-processing pipelines, and specialized AI-based systems designed for narrowly scoped analytical tasks.

What defines a CMI is not the presence of machine learning, but its methodological role. A CMI implements a well-defined operational procedure *once a computational pathway has been selected*. Examples include molecular dynamics engines, econometric modeling platforms, numerical inference libraries, geospatial analysis tools, and domain-specific computational systems embedded in established technical workflows.

Importantly, CMIs do not function as cognitive agents and do not participate in epistemic decision-making. They encode how an operation can be performed, not when, why, or whether it should be executed. Within the CUA framework, CMIs are identified and contextualized during instrumental mapping as part of the epistemic landscape of inquiry, but they are never executed by the system itself. Execution—when required—occurs strictly outside the CUA, under explicit human control, after epistemic convergence has been achieved.

**4.3 Instrumental Awareness as an Epistemic Outcome**

Within the CUA, the identification and organization of UCIs—including CMIs—constitutes a distinct epistemic outcome: instrumental awareness. Instrumental awareness refers to the explicit visibility of what kinds of resources exist to address a problem, what assumptions they encode, and what constraints they impose, prior to any commitment to execution or intervention.

This form of awareness is neither prescriptive nor evaluative. It does not assert feasibility, optimality, or correctness. Instead, it renders the *space of possible inquiry* legible, supporting reflective human judgment about whether, how, and under what conditions further action should occur outside the CUA.

By treating instruments as epistemic resources rather than executable actions, the CUA enables a form of AI-assisted reasoning that is both expansive and disciplined: expansive in its exposure of methodological and institutional possibilities, and disciplined in its refusal to collapse reasoning into autonomous action. This instrumental visibility provides the conceptual foundation for the evaluation metrics introduced in the following section, which quantify not only convergence and alignment, but also the breadth and structure of instrumental awareness achieved during inference.

**5. Methodology and Evaluation Metrics**

The objective of this study is not to assess task accuracy, factual correctness, or downstream utility, but to characterize how different inference organizations shape epistemic behavior during AI-assisted reasoning. Accordingly, the evaluation framework focuses on *process-level properties* of inference rather than on output-level performance.

We conduct a controlled comparative evaluation between two inference paradigms that differ solely in how cognitive labor is allocated and governed during inference: (i) *CUA-orchestrated inference*, which implements Explicit Cognitive Allocation through externally enforced cognitive staging, and

(ii) *baseline LLM inference*, which relies on unconstrained generative continuation without explicit epistemic role separation.

All experiments are designed to isolate the effects of inference organization from model capacity, training data, or prompt formulation.

**5.1 Experimental Design**

We evaluate both inference paradigms under matched execution conditions, using the same underlying language model(s), identical prompts, and fixed sampling parameters. No fine-tuning, prompt optimization, or model-specific heuristics are introduced in either condition.

The evaluation is architecture-centered rather than prompt-centered. Each paradigm is treated as a distinct cognitive organization that transforms a single human intent into a sequence of epistemic artifacts and, ultimately, a convergent synthesis.

To ensure comparability, both paradigms are constrained to the same number of observable reasoning steps. Each step corresponds to a discrete language-model invocation and constitutes an evaluation checkpoint at which epistemic state, alignment, and instrumental visibility can be measured.

The experimental domain selected for evaluation is agricultural systems, chosen for its combination of scientific complexity, institutional constraints, methodological heterogeneity, and strong dependence on instrumental mediation. This domain provides a stringent testbed for assessing epistemic alignment and instrumental awareness without privileging narrow technical optimization.

### 5.2 Inference Paradigms and Cognitive Allocation Regimes

The two inference paradigms compared in this study differ exclusively in how cognitive functions are organized during inference.

To maintain analytical clarity, we distinguish between cognitive stages and evaluation checkpoints:

- Cognitive stages: are abstract epistemic units defined by the AI-assisted workflow (Section 4). They describe *what kind of epistemic work is being performed*.
- Evaluation checkpoints: are experimentally defined observation points corresponding to concrete model invocations, at which metrics are computed.

This distinction allows epistemic organization to be analyzed independently of implementation details.

#### 5.2.1 CUA-Orchestrated Inference

In the CUA condition, inference is conducted through the AI-assisted workflow defined in Section 4. The Cognitive Universal Agent enforces explicit cognitive staging, assigning each stage a well-defined epistemic objective and a bounded non-executive role.

The workflow comprises four conceptual stages:

1. Exploration and framing: which clarifies and stabilizes the initial human intent;
2. Epistemic anchoring and instrumental mapping: which situates the inquiry within existing knowledge and identifies relevant Universal Cognitive Instruments (UCIs) without executing them;

3. Operational design (non-executable): which organizes potential methodological and procedural options;
4. Interpretation and synthesis: which integrates epistemic artifacts into a coherent convergent representation.

For evaluation purposes, these four stages are observed through five execution checkpoints. The additional checkpoint arises from an analytical separation between epistemic synthesis and final narrative realization. This separation is introduced solely to increase measurement resolution and does not constitute an additional cognitive stage. Cognitive organization is therefore externally governed and reproducible, rather than emerging dynamically from conversational interaction or model heuristics. A detailed description of checkpoint implementation and logging is provided in the Supplementary Information.

**5.2.2 Baseline LLM Inference**

In the baseline condition, inference proceeds through chained generative invocations without explicit cognitive role separation. Conceptual framing, epistemic grounding, instrumental suggestion, and synthesis are implicitly interleaved within a single generative process.

Each model invocation constitutes an evaluation checkpoint, but no architectural constraints enforce epistemic boundaries between steps. Convergence is achieved through continued generation rather than through externally defined closure criteria.

To ensure comparability, the baseline condition is constrained to the same number of evaluation checkpoints as the CUA condition. This design choice ensures that observed differences reflect cognitive allocation and workflow organization, rather than interaction length, token budget, or model capacity.

**5.3.1 Length of Workflow to Convergence (LWC)**

Length of Workflow to Convergence (LWC) measures the number of model invocations required to reach epistemic convergence under the experimental protocol.

In the CUA condition, convergence is achieved through a fixed, architecturally defined workflow, yielding an invariant value of LWC. In the baseline LLM condition, convergence requires an additional generative step to achieve closure.

LWC therefore captures structural discipline in inference closure, distinguishing convergence achieved by architectural constraint from convergence achieved through extended generation.

**5.3.2 Semantic Deviation, Epistemic Alignment, and Anchored Expansion**

To characterize epistemic behavior beyond convergence speed, we introduce a coupled set of semantic metrics that distinguish controlled epistemic growth from uncontrolled semantic drift.

The original human prompt is treated as a fixed epistemic anchor $A$, representing the invariant research intent.

**Semantic Deviation Rate (TDS)**

Semantic Deviation Rate (TDS) quantifies how far the epistemic objectives guiding inference move away from the anchor as reasoning unfolds. It is computed as the average semantic distance between the anchor and the declared epistemic objective at each checkpoint.

Low TDS values indicate limited expansion; high values indicate substantial epistemic movement.

**Epistemic Alignment Score (EAS)**

Epistemic Alignment Score (EAS) measures whether semantic expansion remains aligned with the original intent. High EAS values indicate coherent expansion; low values indicate divergence.

**Anchored Epistemic Expansion (AEE)**

Anchored Epistemic Expansion combines deviation and alignment into a single measure:

$$AEE = TDS \times EAS$$

AEE is maximized when inference exhibits substantial expansion while remaining aligned with the epistemic anchor. It distinguishes productive inquiry from semantic drift or trivial reasoning.

**5.4 Instrumental Awareness Metrics**

Beyond convergence and semantic alignment, Explicit Cognitive Allocation is designed to make the instrumental structure of inquiry explicit. To quantify this effect, we introduce metrics that measure instrumental visibility rather than instrumental execution.

**5.4.1 Instrumental Coverage Index (ICI)**

The Instrumental Coverage Index (ICI) measures the number of distinct classes of Universal Cognitive Instruments (UCIs) that are explicitly identified during inference.

$$ICI = |\{\text{distinct UCI classes identified}\}|$$

An instrument contributes to ICI when it is explicitly surfaced as relevant to addressing the inquiry. No feasibility assessment or execution plan is required.

ICI therefore measures epistemic breadth, not correctness or optimality.

**5.4.2 Instrumental Exploration Score (IES)**

Instrumental Exploration Score (IES) captures how systematically an inference process explores the instrumental landscape, accounting for both coverage and structural differentiation among instruments.

While ICI measures *how many* classes are surfaced, IES measures *how coherently* they are articulated and contextualized relative to the inquiry.

Formal definition and computation details for IES are provided in the Supplementary Information.

**5.5 Summary**

Together, these metrics operationalize the core claims of Explicit Cognitive Allocation. They enable systematic comparison between inference paradigms based on:

- how convergence is achieved,
- how semantic expansion is governed, and
- how explicitly the instrumental landscape of inquiry is rendered visible.

In the following section, we report empirical results obtained under this framework and analyze how explicit cognitive allocation reshapes inference behavior relative to baseline LLM inference.

**6. Results.**

Results are organized according to the epistemic structure of AI-assisted reasoning rather than conventional performance benchmarks. We first analyze semantic stability and alignment, followed by epistemic traceability, instrumental awareness, and aggregate token usage. This ordering reflects a dependency among metrics: preservation of the original epistemic intent is a prerequisite for interpreting differences in efficiency or instrumental scope.

Length of Workflow to Convergence (LWC_eff) is treated as a structural property fixed by architectural design rather than as an empirical outcome. Under the experimental protocol, CUA-orchestrated inference converges through four epistemic stages (LWC_eff = 4), whereas baseline LLM inference requires five conversational steps (LWC_eff = 5). This difference reflects distinct cognitive organizations of inference, not differences in speed or efficiency.

Accordingly, LWC_eff functions as a controlled boundary condition for all subsequent analyses. All results are based on matched-step comparisons, ensuring that observed differences arise from cognitive and instrumental organization rather than from model capacity, prompting strategy, or interaction length.

**6.1 Case study: "Evalúa prácticas agroecológicas".**

The empirical behavior of the Semantic Deviation Rate (TDS) is illustrated through the prompt *"Evalúa prácticas agroecológicas"*, executed under two distinct cognitive paradigms: an explicit-cognition architecture (CUA) and a baseline large language model operating under implicit cognition (LLM). In the LLM condition, inference proceeds as a fixed reflective loop of five reported steps. However, these steps do not correspond to independently declared epistemic objectives. Instead, the objectives $O_i$ required for computing TDS are approximated post hoc from the model's inferred continuations, capturing implicit goal evolution within a single latent inference process. As a consequence, the execution-level TDS reflects the average semantic expansion of the inquiry relative to the original anchor, but without any architectural mechanism for identifying, localizing, or constraining that expansion during reasoning.

Empirically, this results in relatively high and tightly clustered TDS values (Fig. 1a), indicating substantial epistemic expansion that remains narratively coherent but is not explicitly regulated. In this setting, low variability in TDS should not be interpreted as epistemic stability, but rather as the model's capacity to smoothly absorb semantic deviation within a unified narrative space. TDS here functions as a descriptive, post hoc measure of expansion, not as an operational control signal.

By contrast, the CUA implements an explicit cognitive workflow in which epistemic objectives $O_i$ are declared, logged, and associated with functionally differentiated stages, including conceptual exploration, instrumental mapping, and interpretive synthesis. This enables the computation of stage-specific $TDS_i$, allowing semantic expansion to be detected and evaluated at the point of emergence. As a result, execution-level TDS exhibits slightly greater variability, but remains bounded, reflecting controlled epistemic expansion rather than uncontrolled drift.

Crucially, this architectural distinction determines the epistemic meaning of TDS. While both paradigms allow computation of TDS as defined above, only the CUA endows the metric with interpretive significance by coupling semantic deviation to explicit epistemic objectives. In the LLM baseline, TDS captures how far the inquiry moves; in the CUA, TDS captures how and where that movement occurs. This distinction generalizes across the full set of twenty AI-generated prompts spanning agroecological, environmental, and sustainability-related domains, as summarized at the population level in Fig. 1.

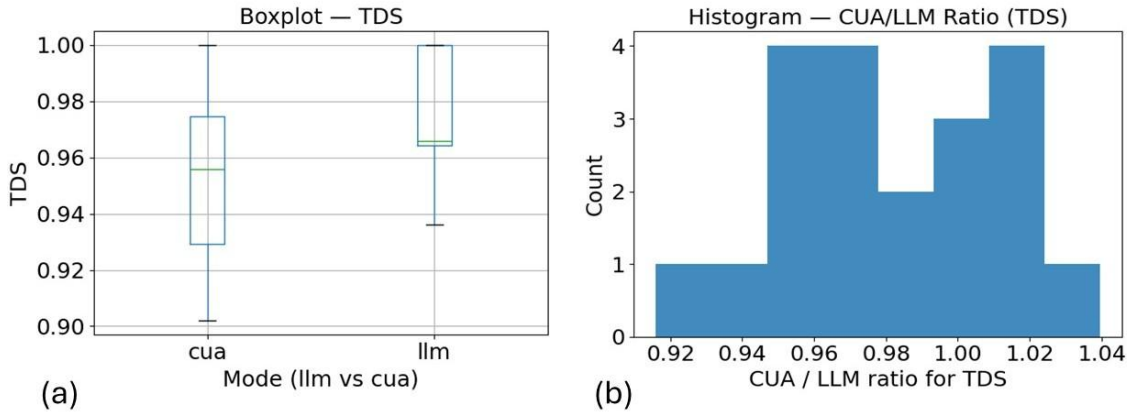

**Figure 1.** (a) Boxplot comparison of Task Deviation Score (TDS) across matched inference executions under baseline LLM inference and CUA-orchestrated inference. The x-axis indicates the inference paradigm (LLM vs. CUA), and the y-axis reports TDS, defined as the mean semantic distance between the original epistemic anchor and the epistemic content generated during inference. Boxes represent the interquartile range (25th–75th percentiles), the central line denotes the median, and whiskers extend to 1.5×IQR. (b) Distribution of the CUA/LLM TDS ratio computed at the execution level. Values near unity indicate comparable semantic deviation across paradigms, whereas deviations above or below one reflect relative increases or decreases in thematic drift under CUA with respect to the baseline LLM.

Figure 1(a) shows that the execution-level Semantic Deviation Rate (TDS) obtained under CUA-orchestrated inference and baseline LLM inference occupies a similar numerical range, with overlapping interquartile intervals and comparable medians. This indicates that, at an aggregate level, both paradigms exhibit a comparable degree of epistemic expansion relative to the original anchor. Figure 1(b) reinforces this observation: the distribution of the CUA/LLM TDS ratio remains tightly clustered around unity, with only modest deviations above or below one. Quantitatively, this suggests that the *amount* of semantic expansion produced by the two systems is broadly similar across prompts.

However, this numerical proximity masks a critical architectural distinction. In the baseline LLM condition, TDS reflects expansion emerging from a fully collapsed latent inference process, in which epistemic objectives are inferred retrospectively and integrated narratively. As illustrated by the LLM-generated synthesis, conceptual broadening—e.g., the shift from evaluative assessment to normative claims about sustainability, resilience, and socio-economic transformation—is smoothly absorbed into the narrative without explicit signaling of where or why the expansion occurs. In this setting, TDS functions as a descriptive measure of semantic movement, but does not distinguish controlled inquiry from unregulated drift.

By contrast, under CUA orchestration, similar levels of TDS correspond to epistemic expansion that is explicitly governed. Objectives are declared, localized, and constrained at the level of workflow stages, allowing semantic deviation to be interpreted as *intentional exploration* rather than implicit narrative drift. Thus, while Figures 1a and 1b show that TDS magnitudes alone do not separate the two paradigms, they also demonstrate why TDS by itself is insufficient: identical or near-identical values can arise from fundamentally different cognitive organizations—semantic continuity in the LLM versus governed epistemic expansion in CUA.

While the Semantic Deviation Rate (TDS) provides a quantitative measure of how far an inference process departs from its original epistemic anchor, it does not, by itself, distinguish between productive expansion and uncontrolled semantic drift. As shown in Fig. 1, both baseline LLM inference and CUA-orchestrated inference exhibit comparable levels of semantic deviation, suggesting that distance from the initial intent alone is insufficient to characterize epistemic quality. Similar TDS values may arise either from coherent exploration governed by explicit epistemic constraints or from unguided narrative continuation under implicit inference.

To address this limitation, we introduce the Epistemic Alignment Score (EAS), which measures the semantic coherence between the original epistemic anchor and the final synthesized output. Unlike TDS, which aggregates deviation across intermediate objectives, EAS is evaluated at the level of the final synthesis and captures whether the resulting expansion remains conceptually aligned with the initial intent. In the CUA architecture, this final synthesis emerges from an explicitly governed workflow in which epistemic roles, cognitive functions, and instrumental mappings are declared and constrained. In contrast, under baseline LLM inference, alignment is an emergent property of conversational continuation rather than an explicitly regulated outcome.

Because neither deviation nor alignment alone is sufficient to characterize epistemic performance, we define Anchored Epistemic Expansion (AEE) as the product of TDS and EAS. This composite metric operationalizes the distinction between mere semantic movement and epistemically productive expansion: high AEE values correspond to inference processes that explore beyond the initial framing while preserving conceptual coherence with the original epistemic anchor. Figure 2 summarizes the behavior of EAS, AEE, and their relative ratios across inference paradigms, highlighting systematic differences that remain invisible when considering TDS in isolation.

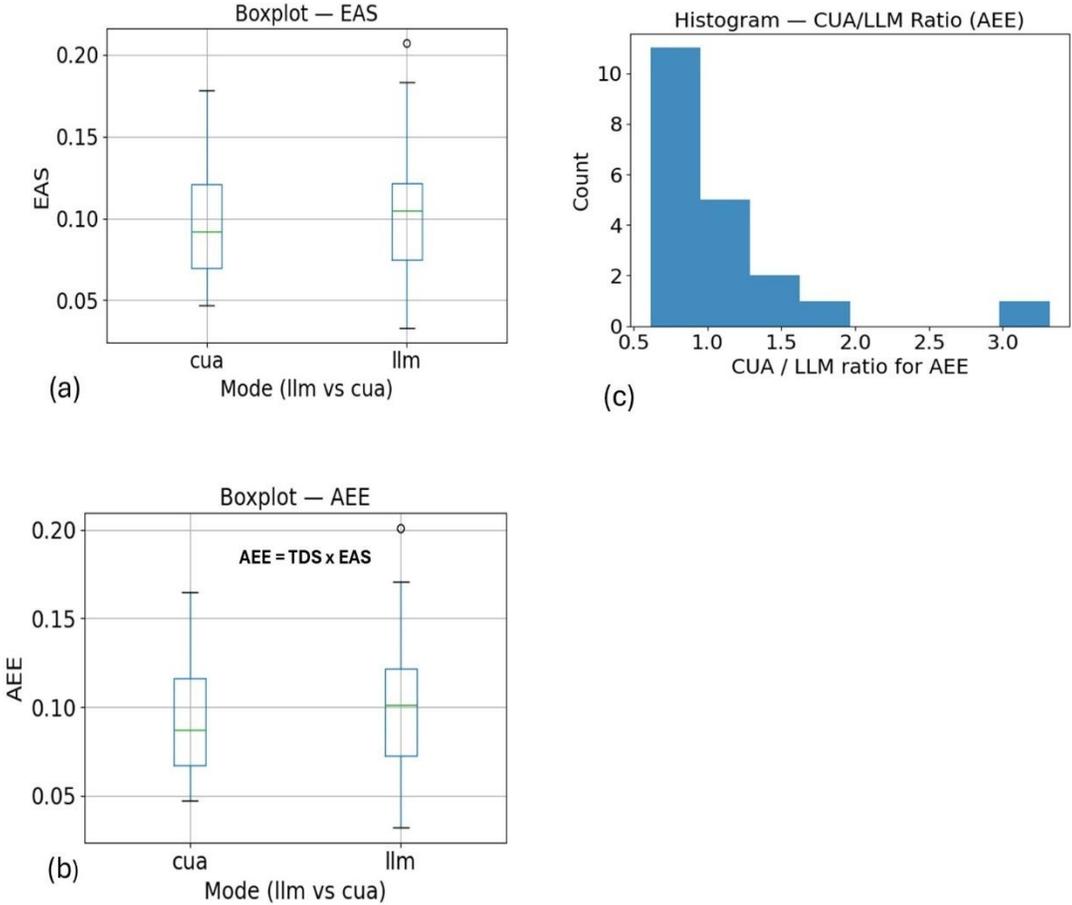

Figura2. (a) Boxplot comparison of the Epistemic Alignment Score (EAS) across matched inference executions under baseline LLM inference and CUA-orchestrated inference. The x-axis indicates the inference paradigm (LLM vs. CUA), and the y-axis reports EAS, computed as the semantic similarity between the original epistemic anchor and the final synthesized output. Boxes represent the interquartile range (25th–75th percentiles), the central line denotes the median, and whiskers extend to 1.5×IQR. (b) Boxplot comparison of Anchored Epistemic Expansion (AEE) across the same matched executions, where AEE is defined as the product of the Semantic Deviation Rate (TDS) and EAS (AEE = TDS × EAS). Axes and boxplot conventions follow panel (a). (c) Distribution of the execution-level CUA/LLM ratio for AEE. Values near unity indicate comparable levels of anchored epistemic expansion across paradigms, whereas deviations above or below one reflect relative differences between CUA-orchestrated and baseline LLM inference.

Figure 2 reports the distribution of the Epistemic Alignment Score (EAS) and Anchored Epistemic Expansion (AEE) across matched inference executions under baseline LLM inference and CUA-orchestrated inference. Across executions, both paradigms exhibit EAS values concentrated in a comparable range, indicating that final outputs remain semantically related to the original epistemic anchor in both conditions. Median EAS values vary across prompts, with neither paradigm showing systematic dominance, and isolated high-EAS cases appearing in both LLM and CUA executions.

AEE values closely track the behavior of EAS, as expected from its definition as the product of TDS and EAS (AEE = TDS × EAS). Because TDS values are consistently high and close to unity in both paradigms (typically in the range 0.90–1.00), variation in AEE is primarily driven by variation in EAS rather than by differences in semantic deviation magnitude. Consequently, AEE largely reflects differences in alignment of expanded content with the original epistemic anchor, rather than differences in the extent of expansion itself.

The execution-level comparison shows that AEE values for CUA and LLM are of similar order of magnitude, with prompt-dependent alternation in which paradigm attains higher AEE. This pattern is reflected in panel (c), where the distribution of the CUA/LLM AEE ratio is concentrated between approximately 0.5 and 1.0, with a right-skewed tail and a small number of executions exceeding unity. Ratios near one indicate executions where both paradigms achieve comparable anchored epistemic expansion, whereas ratios below one correspond to cases where higher alignment-weighted expansion is observed under the baseline LLM condition. However, these numerical comparisons should not be interpreted as evidence that baseline LLM inference is intrinsically superior to CUA-orchestrated inference, as the two paradigms implement fundamentally different inference architectures and epistemic control mechanisms.

While Figures 1 and 2 establish that baseline LLM inference and CUA-orchestrated inference exhibit comparable levels of semantic deviation and alignment at the level of final outputs, they do not address a critical dimension of reasoning: whether the means of inquiry themselves are made explicit. Figure 3 examines this instrumental dimension by quantifying how inference paradigms differ in their ability to surface, organize, and externalize the instruments through which abstract questions become investigable.

Figure 3a reports the distribution of the normalized Instrumental Coverage Index ($ICI_n$), which measures the number of distinct classes of instruments explicitly identified during inference. Under the baseline LLM condition, $ICI_n$ remains near zero across all executions, indicating that instrumental references—although occasionally present in narrative form—are not systematically identified, categorized, or stabilized as epistemic objects. In contrast, CUA-orchestrated inference consistently exhibits maximal or near-maximal $ICI_n$ values, reflecting the explicit declaration of multiple instrument classes spanning computational, experimental, organizational, regulatory, and economic domains.

Importantly, this difference does not arise from superior domain knowledge or richer content generation in the CUA. As illustrated by the baseline LLM outputs, similar instruments, institutions, and technologies may appear implicitly within narrative responses. However, in the absence of an explicit instrumental mapping stage, these references remain entangled within prose, rendering them inaccessible to measurement, comparison, or audit. ICI thus captures not the presence of instruments in text, but their epistemic externalization as structured and countable artifacts.

Figure 3b further examines the depth of instrumental characterization using the Instrumental Exploration Score (IES). Baseline LLM inference again exhibits minimal values, indicating that instruments—when mentioned—are rarely contextualized in terms of function, constraints, or relevance to the original epistemic objective. By contrast, CUA-orchestrated inference consistently yields high IES values, reflecting systematic characterization of instruments across multiple attributes, including purpose, scope, limitations, and institutional embedding. This separation between coverage (ICI) and depth (IES) highlights that instrumental awareness is not merely a function of enumeration, but of structured epistemic treatment.

Figure 3c provides a concrete illustration of this distinction by showing a structured technical synthesis produced by the CUA. The synthesis explicitly records workflow stages, declared Universal Cognitive Instrument (UCI) classes, and metric availability, and is preserved as a non-user-facing technical artifact. This artifact is not an explanatory afterthought, but a byproduct of the inference process itself, enabling auditability, replicability, and epistemic inspection at the level of reasoning structure rather than narrative output.

Taken together, Figure 3 demonstrates that instrumental awareness constitutes an orthogonal dimension of inference behavior that is not captured by semantic deviation, alignment, or expansion metrics alone. While baseline LLM inference can produce fluent and informative syntheses, it collapses instrumental reasoning into narrative generation, leaving no inspectable trace of how methods, tools, and institutions were selected or prioritized. The CUA, by contrast, externalizes this dimension by design, transforming instruments from implicit narrative elements into explicit epistemic objects.

This distinction clarifies why increased token expenditure under CUA orchestration (as reported in the operational cost analysis) should not be interpreted as inefficiency. The additional computational cost reflects the deliberate construction of auditable epistemic structure rather than redundant generation. The value of the CUA therefore lies not in producing "better answers," but in enabling forms of epistemic accountability and methodological transparency that are unattainable under monolithic LLM-based inference.

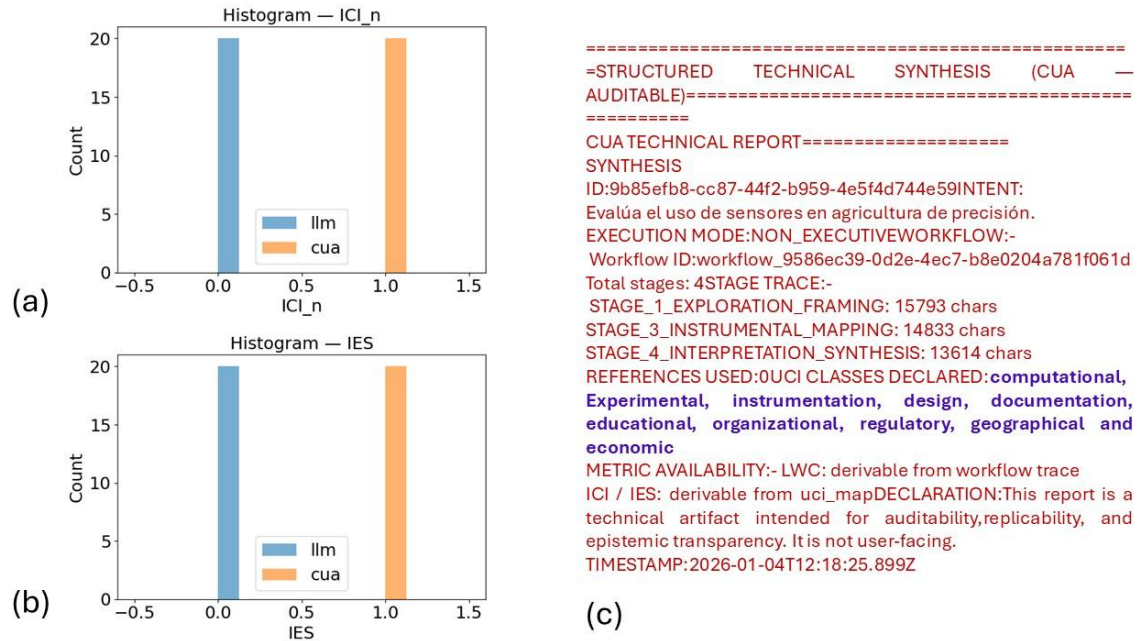

Figure 3 (a) Distribution of the normalized Instrumental Coverage Index (ICI $_n$) across matched inference executions, quantifying the number of distinct classes of instruments explicitly identified during reasoning. Baseline LLM inference exhibits near-zero instrumental coverage, whereas CUA-orchestrated inference consistently exposes a broad and stable instrumental space. (b) Distribution of the Instrumental Exploration Score (IES), measuring the depth with which identified instruments are characterized and contextualized. Instrumental exploration remains minimal under baseline LLM inference and is markedly higher under CUA orchestration, reflecting explicit instrumental mapping.(c) Example of a structured technical synthesis generated by the CUA, showing explicit declaration of workflow stages, Universal Cognitive Instrument (UCI) classes, and metric availability. The synthesis is preserved as an auditable, machine-traceable technical artifact intended for epistemic transparency and replicability.

**Operational cost, auditability, and architectural implications**

To clarify the operational implications of explicit cognitive orchestration, we compared total token usage across both inference paradigms. Under the experimental protocol, CUA-orchestrated inference consumed, on average, approximately five times more tokens than baseline LLM inference. This difference reflects the explicit externalization of cognitive structure—epistemic staging, objective declaration, and instrumental mapping—rather than differences in model capacity or parameterization. Token usage therefore captures the operational cost of architectural transparency, not epistemic value or output quality.

This distinction is central when considering auditability as a structural property of inference. Large language models generate coherent outputs through a single collapsed generative process in which conceptual framing, retrieval, inference, and synthesis are internally conflated. As a result, LLM-based inference can be evaluated only at the level of outcomes.

Post hoc explanations constitute new generations rather than recoverable traces of the original reasoning process and therefore do not enable genuine auditability.

The Cognitive Universal Agent introduces auditability by design. By externalizing reasoning into explicitly staged epistemic artifacts—problem framing, epistemic anchoring, instrumental mapping, and interpretive synthesis—each preserved as a traceable object, the CUA enables inspection, comparison, and revision of epistemic decisions independently of the final narrative realization. In this sense, higher token consumption reflects the cost of making reasoning explicit, governable, and accountable.

Beyond experimental evaluation, this architectural property has direct implications for real-world deployment contexts in which decisions must be justified, reviewed, or contested. Domains such as insurance underwriting, regulatory compliance, legal and consulting practices, and high-stakes organizational decision-making—particularly across the United States, Europe, and other regulated environments—require not only plausible answers, but traceable reasoning structures and explicit methodological grounding. In such settings, CUA-like architectures are best understood as complements to generative models, mediating between statistical language generation and institutional requirements for accountability, documentation, and review.

Taken together, these results position explicit cognitive and instrumental allocation as a foundational architectural principle for AI-assisted reasoning. The contribution of the CUA does not lie in semantic superiority, but in reorganizing cognitive labor into auditable, stage-separated processes that resist the epistemic opacity of monolithic inference. This reframes efficiency not as output brevity, but as structured convergence toward epistemic closure. Future work should therefore focus on hybrid architectures that integrate generative fluency with explicit epistemic governance, and on execution-level instrumentation to quantify computational, organizational, and regulatory trade-offs as emergent properties of cognitive architecture rather than model internals.

**References.**

1  Floridi, L. & Chiriatti, M. GPT-3: Its nature, scope, limits, and consequences. *Minds Mach*. 30, 681–694 (2020).

2  Binz, M. & Schulz, E. Using cognitive psychology to understand GPT-3. *Proc. Natl Acad. Sci. USA* 120, e2218523120 (2023).

3  Bender, E. M., Gebru, T., McMillan-Major, A. & Shmitchell, S. On the dangers of stochastic parrots: Can language models be too big? In *Proc. ACM FAccT* 610–623 (2021).

4  Russell, S. *Human Compatible: Artificial Intelligence and the Problem of Control*. (Viking, 2019).

5  Wei, J. *et al*. Chain-of-thought prompting elicits reasoning in large language models. *Adv. Neural Inf. Process. Syst.* 35, 24824–24837 (2022).

6  Yao, S. *et al*. Tree of thoughts: Deliberate problem solving with large language models. *Adv. Neural Inf. Process. Syst.* 36 (2024).

7  Besta, M. *et al*. Graph of thoughts: Solving elaborate problems with large language models. *Proc. AAAI Conf. Artif. Intell.* 38, 17682–17690 (2024).

8  Deng, X. *et al*. Optimizing reasoning traces in large language models. *Adv. Neural Inf. Process. Syst.* 36 (2024).

9  Lewis, P. *et al*. Retrieval-augmented generation for knowledge-intensive NLP tasks. *Adv. Neural Inf. Process. Syst.* 33, 9459–9471 (2020).

10  Schick, T. *et al*. Toolformer: Language models can teach themselves to use tools. *Adv. Neural Inf. Process. Syst.* 36 (2024).

11  Meskó, B. & Topol, E. J. The imperative for regulatory oversight of large language models in healthcare. *npj Digit. Med.* 6, 120 (2023).

12  Hutchins, E. *Cognition in the Wild*. (MIT Press, 1995).

13  Giere, R. N. Scientific cognition as distributed cognition. In *The Cognitive Basis of Science* 285–299 (Cambridge Univ. Press, 2002).

14  Goldman, A. I. *Knowledge in a Social World*. (Oxford Univ. Press, 1999).